\documentclass[11pt]{article}
\usepackage{lipsum}  
\usepackage{hyperref} 
\usepackage{graphicx} 
\usepackage{natbib} 
\usepackage{multirow}
\usepackage{makecell}
\usepackage{graphicx} 
\usepackage{array}
\usepackage{lscape}
\usepackage{float}
\usepackage{multirow} 
\usepackage{booktabs} 
\usepackage{geometry} 
\usepackage{caption} 
\usepackage{pdflscape} 
\usepackage{fancyhdr}
\usepackage{adjustbox}
\usepackage{longtable} 
\usepackage{longtable}
\title{A Cross-Validation Study of Turkish Sentiment Analysis Datasets and Tools}

\author{
    \begin{tabular}{c c}
        Şevval Çakıcı & Dilara Karaduman \\
        \href{mailto:sevval.cakici@ozu.edu.tr}{\texttt{sevval.cakici@ozu.edu.tr}} & \href{mailto:dkaraduman17@ku.edu.tr}{\texttt{dkaraduman17@ku.edu.tr}} \\
        Ozyegin University & Koc University \\
        & \\
        Mehmet Akif Çırlan & Ali Hürriyetoğlu \\
        \href{mailto:mcirlan19@ku.edu.tr}{\texttt{mcirlan19@ku.edu.tr}} & \href{mailto:ali.hurriyetoglu@wur.nl}{\texttt{ali.hurriyetoglu@wur.nl}} \\
        Koc University & Wageningen Food Safety Research \\
    \end{tabular}
}
   
\date{\today}

\begin{document}
\maketitle

\begin{abstract}
In recent years, sentiment analysis has gained increasing significance, prompting researchers to explore datasets in various languages, including Turkish. However, the limited availability of Turkish datasets has led to their multifaceted usage in different studies, yielding diverse outcomes. To overcome this challenge, a rigorous review was conducted of research articles published between 2012 and 2022. 31 studies were listed, and 23 Turkish datasets obtained from publicly available sources and email requests used in these studies were collected. We labeled these 31 studies using a taxonomy. We provide a map of sentiment analysis datasets according to this taxonomy in Turkish over 10 years. Moreover, we run state-of-the-art sentiment analysis tools on these datasets and analyzed performance across popular Turkish sentiment datasets. We observed that the performance of the sentiment analysis tools significantly depends on the characteristics of the target text. Our study fosters a more nuanced understanding of sentiment analysis in the Turkish language. 

\end{abstract}


\section{Introduction}
The world is closely associated with people's feelings and behaviors, ranging from research on societal trends to applications in product marketing. In recent years, the need for a systematization of understanding emotions and behaviors has become even more pronounced. Typically, scholars frame emotion as a mental experience, and the necessity to define and classify emotions has led to models such as Plutchik's wheel of emotions \citep{Plutchik1980}. \citet{Ekman1992} focused on six basic emotions, often referred to as Ekman categories in the text processing literature. According to \citet{Cabanac2002}, when there is no standard definition in the literature, researchers usually define emotion concerning a list: anger, disgust, fear, joy, sadness, and surprise. These categorizations are foundational in many fields, especially in sentiment analysis and text processing. 

With the development of technology, data has started to be recorded in the digital world more than ever in the history of humanity, with platforms like Facebook, Twitter, and Instagram serving as prime platforms of this digital data explosion. The vast amounts of user-generated online content provide an unprecedented opportunity to study human behavior on a large scale. Since the recorded data is accessible to everyone, it also plays a role in emerging new behaviors. For example, the expected behavior of the 21st-century consumer includes looking at the opinions of others online before making a decision \citep{Yayli2012}. Likewise, companies analyze consumer behavior and opinions similarly before making decisions and launching a product. Such big data, which includes people's thoughts, feelings, opinions, and practices, also serves as a source for scientific research. 

Sentiment analysis, which arises as a result of the need to analyze behavior, evaluate people's attitudes, and orientations, and classify their emotions, is typically concerned with determining the overall positive or negative stance expressed in text \citep{Devika2016}. For many practical applications, understanding this general sentiment is more critical than identifying specific emotions. Sentiment analysis, also known as opinion mining, is the computational study of people's sentiments, feelings, views, and attitudes toward a particular topic, situation, event, or product \citep{Zhang2018}. This approach enables governments to notice mass referrals and helps companies analyze the behavior and views of their target audience. According to \citet{Zhang2018}, these practices and industrial interests stimulated high motivations for sentiment analysis research.

Sentiment analysis studies are conducted in many languages, as \citet{Abbasi2008} indicated. Among these languages, Turkish ranked fourth in common languages for web content in January 2021, according to `The Most Used Languages on the Internet – Visual Capitalist Licensing' (n.d.)\citep{VisualCapitalist2022}. As of July 2022, Internet World Stats (2022) reports Turkey's population as 85,839,173, with 72,500,000 internet users, underscoring the statistical significance of the Turkish language on the internet \citep{InternetWorldStats2022}. However, the limited availability of datasets in some commonly used languages in scientific environments leads to using the same or similar datasets in different studies. For instance, examining Turkish sentiment analysis studies published between 2012 and 2022 reveals that researchers created and used only 23 datasets in 31 studies. Most of these data sets contain less than ten thousand samples. These small-scale datasets are not obtained from a wide variety of sources. The most frequently recurring dataset is the one created by \citet{Demirtas2013}, utilized not only in their studies but also in works by \citet{Parlar2016}, \citet{Mulki2018} and \citet{Alqaraleh2021}.

One aspect needing more focus in Turkish sentiment analysis publications is the comparative presentation of these studies. To address this, we conducted an investigation on research papers on `sentiment analysis' and `Turkish dataset' on Scopus between 2012 and 2022 to address this challenge. Subsequently, 23 datasets from 31 studies, obtained through publicly available sources and email requests from authors of respective studies, were compiled and organized in a GitHub repository\footnote{\url{https://github.com/sevvalckc/Turkish-SAD}}. In this study, we assign labels to these publications based on the sentiment analysis taxonomy classification \citep{Rodrigues2018} to provide a comprehensive perspective. The goal is to offer a valuable resource for future sentiment analysis research in Turkish by classifying and comparing these studies. 

Furthermore, the methodology employed aligns with the principles of a Systematic Literature Review (SLR), aiming to identify relevant primary studies, extract the required information regarding the research questions (RQs), and synthesize the information to respond to these RQs. An SLR follows a well-defined methodology and assesses the literature in an unbiased and repeatable way \citep{Kitchenham2007}.

We generate predictions using the multilingual XLM-T model \citep{Barbieri2021}, trained on 198 million tweets and fine-tuned for sentiment analysis, as well as the BERTurk model enhanced with the Bounti dataset \citep{Koksal2021}, the Turkish Sentiment Analysis model -briefly TSAM- \citep{Tasar2022}, and TurkishBERTweet developed by ViralLab \citep{Najafi2023} applied to four datasets that were recurrent in more than one publication among the 31 mentioned above. Another purpose of this study is to compare the best-reported model predictions from the studies and the predictions of the models. For this purpose, it is desired to shed light on the reasons for the low accuracy scores and to contribute to the cross-domain classification literature with the compiled information~\cite{Hurriyetoglu+2021}.

Our study is designed to achieve two primary objectives. Firstly, it aims to provide a comprehensive and detailed overview of sentiment analysis in the Turkish language, offering a systematic classification and thorough comparison of existing research between 2012 and 2022. Secondly, the study focuses on comparing and discussing Turkish sentiment analysis tools in the context of dataset characteristics. By undertaking these objectives, we seek to contribute valuable insights to the field of Turkish sentiment analysis, facilitating a deeper understanding of the methodologies employed, dataset variations, and the overall landscape of sentiment analysis research in the Turkish language.

The upcoming parts of this article are organized as follows. In section~\ref{sec:related_work}, the related studies are compiled as both sentiment analysis and cross-dataset classification studies. In section~\ref{sec:datasets}, specific information about the datasets is given, and the datasets are compared in terms of similarities. Next, Section~\ref{sec:methodology} describes the methodology, including the models used and the sentiment analysis process for Turkish datasets. Section~\ref{sec:results} presents the results of the selected models on various datasets. Section~\ref{sec:discussion} provides a discussion of the findings, along with limitations and implications for future research. Finally, section~\ref{sec:conclusions_and_future_work} concludes the study and suggests areas for further exploration in Turkish sentiment analysis. 

\section{Related Work}
\label{sec:related_work}
Our research focuses on sentiment analysis, a field predominantly concentrated on English language studies. However, we are specifically focusing on the Turkish language. Acknowledging this methodology, our study aims to address this gap by systematically compiling and examining existing sentiment analysis research and datasets in Turkish. This effort is crucial for advancing the understanding and development of sentiment analysis research, tools, and methodologies for the Turkish language, thereby contributing to the improvement and enrichment of the Turkish NLP field.

A comprehensive study by \citet{altinel2024turkish} reviews the methods used in Turkish sentiment analysis, categorizing them into three main approaches: dictionary-based, machine learning-based, and hybrid. Their analysis highlights the strengths and weaknesses of each method, particularly emphasizing that machine learning techniques are better suited to Turkish’s complex linguistic structure. Our study, by contrast, examines 31 studies through a taxonomic classification, analyzing 23 datasets in depth. We further compare the performance of four different deep learning models fine-tuned for sentiment analysis across these datasets, revealing the crucial impact of data diversity on model accuracy. This comparative approach adds a broader perspective, focusing on datasets and taxonomy, and highlights the significant role of data preparation in sentiment analysis methods.

Similarly, \citet{fehle2021lexicon}'s study systematically examines the methods of sentiment analysis for German texts and is a review. The study evaluates 19 existing sentiment dictionaries and 20 sentiment annotated corpora for German texts. This comprehensive review analyzes how the dictionaries and datasets perform in different domains, while also detailing the impact of preprocessing steps (stemming, lemmatization, etc.) used before sentiment analysis on performance. In this respect, the study provides a broad map of the methods used in sentiment analysis specific to the German language, similar to the literature reviews of sentiment analysis in Turkish.

Also, \citet{oueslati2020review} provide a detailed review of sentiment analysis research in Arabic. They examine different methods and datasets, focusing on machine learning and deep learning techniques. The study highlights challenges and limitations unique to Arabic sentiment analysis, such as its complex morphology. By pointing out gaps in the existing literature, the authors suggest directions for future research, including developing better datasets and models. Given the similarities in linguistic complexities, their review is a valuable resource for researchers working on Turkish sentiment analysis and dataset development.

\citet{Rodrigues2018} contribute a detailed taxonomy of sentiment analysis studies. We utilized it to further enhance our study. This taxonomy, which is provided in Table~\ref{tabletaxonomy}, is a comprehensive framework that systematically classifies sentiment analysis studies based on critical considerations. It addresses key aspects such as input structure, target class, opinion holder, analysis target, features analyzed in data, data source, application domain, and the approach of lexicon-based, concept-based, and machine learning-based methods. We used the taxonomy presented in Table~\ref{tabletaxonomy} to classify Turkish sentiment analysis studies and datasets in a specific order and technique.

\captionsetup[table]{labelformat=default}
\captionsetup[table]{labelsep=space}

\begin{table}[h!]
    \begin{center}
    \caption{A Taxonomy of Sentiment Analysis Classification from Rodrigues (2018).}
    \label{tabletaxonomy}
    \begin{tabular}{ c c c c c c c }
        \hline
        \multirow{26}{*}{\makecell[c]{Sentiment\\Analysis}} 
        & \multirow{12}{*}{Problem} 
        & \multirow{2}{*}{\makecell[c]{Opinion \\Detection}} 
        & Regular & & & O1 \\ \cline{4-7}
        
        &  &  & Comparative & & & O2 \\ \cline{3-7}
        
        &  & \multirow{2}{*}{\makecell[c]{Entity \\Identification}} 
        & \multirow{2}{*}{Opinion Holder} & Text Author & & E1 \\ \cline{5-7}
        
        &  &  & & Third Person & & E2 \\ \cline{4-7}
        
        &  & & Target Analysis
        & Person & &  E3 \\ \cline{5-7}
        
        &  &  & & Object & &  E4 \\ \cline{5-7}
        
        &  &  & & Organization & &  E5 \\ \cline{3-7}
        
        &  & \multirow{3}{*}{\makecell[c]{Target \\Class}} 
        & Subjectivity & & & T1 \\ \cline{4-7}
        
        &  &  & Polarity & & & T2 \\ \cline{4-7}
        
        &  &  & Emotion & & & T3 \\ \cline{3-7}
        
        &  & \multirow{2}{*}{\makecell[c]{Data\\Specificity}} 
        & Language & & & D1 \\ \cline{4-7}
        
        &  &  & Application Domain & & & D2 \\ \cline{2-7}
        
        & \multirow{6}{*}{Method} 
        & \multirow{6}{*}{\makecell[c]{Granularity \\Analysis}} 
        & Phrase & & & G1 \\ \cline{4-7}
        
        &  &  & Sentence & & & G2 \\ \cline{4-7}
        
        &  &  & Aspect & & & G3 \\ \cline{4-7}
        
        &  &  & Paragraph & & & G4 \\ \cline{4-7}
        
        &  &  & Word & & & G5 \\ \cline{4-7}
        
        &  &  & Document & & & G6 \\ \cline{3-7}
        
        &  & \multirow{3}{*}{Approach} 
        & \multirow{3}{*}{Lexicon Based} & Dictionary Based & & A1 \\ \cline{5-7}

        &  &  & & \multirow{2}{*}{Corpus Based} & Statistical & A2 \\ \cline{6-7}

        &  &  & & & Semantic & A3 \\ \cline{4-7}
        
        &  &  & \multirow{3}{*}{Concept Based} & Ontology & & A4 \\ \cline{5-7}

        &  &  & & Semantic Network & & A5 \\ \cline{4-7}
        
        &  &  & \multirow{3}{*}{\makecell[c]{Machine\\Learning Based}} & Unsupervised & & A6 \\ \cline{5-7}

        &  &  & & Supervised & & A7 \\ \cline{5-7}

        &  &  & & Semi-Supervised & & A8 \\ \cline{1-7}
        
    \end{tabular}
\end{center}
\end{table}

Utilizing this taxonomy is invaluable for researchers in the field, providing a systematic approach to classify studies and datasets. In Table~\ref{tabletaxonomy}, we present the taxonomy derived from \citet{Rodrigues2018}, outlining the categories for analysis. Each category captures essential dimensions of sentiment analysis, guiding researchers to examine studies comprehensively and fostering a standardized approach to classification. This taxonomy serves as a compass, aiding in the organization and understanding of the diverse landscape of sentiment analysis studies in the Turkish language. For instance, the ``Turkish Multidomain Products Reviews" dataset, one of the datasets used in \citet{Demirtas2013} study, includes positive and negative comments received from Hepsiburada.com. This dataset is an ideal example for applying \citet{Rodrigues2018}'s taxonomy. This dataset includes the detection of regular opinions expressed in texts such as ``çok iyi sese sahip bir olan bir kulaklık. alalı neredeyse 1 yıl oldu hiç bir sorunum yok. verilen paraya değer." (EN: A headphone with a very good sound. I bought it almost a year ago and I have no problems. It is worth the money.), and this sentence expresses a positive emotion. In applying the taxonomy specifically to the dataset used in \citet{Demirtas2013} study, the label 'Regular' (O1) is assigned under Opinion Detection because the reviews reflect straightforward opinions about products, rather than comparisons. For Opinion Holder, 'Text Author' (E1) is selected as the opinions are written directly by customers, and 'Third Person' (E2) is included for instances where reviewers refer to others' experiences (e.g., family or friends). Under Target Analysis, 'Product' (E4) is used because the reviews are about specific products such as headphones, while 'Organization' (E5) applies as the reviews often implicitly evaluate Hepsiburada, the platform providing the products. For Target Class, 'Polarity' (T2) is assigned since the dataset classifies reviews into positive or negative sentiments, a clear binary classification. Under Data Specificity, 'Language' (D1) is selected because the dataset is in Turkish, and 'Application Domain' (D2) reflects the focus on e-commerce, specifically product reviews. Regarding Granularity, 'Sentence' (G2) is chosen because the sentiment analysis is performed at the sentence level to capture the opinions expressed in each statement. Finally, 'Supervised Learning' (A7) under Approach is selected because labeled data was used to train machine learning models for sentiment classification. These labels, applied specifically to the Turkish Multidomain Products Reviews dataset in \citet{Demirtas2013} study, systematically define its characteristics and enable structured comparisons within the field of sentiment analysis. These steps were continued and a label was assigned for each study as presented in Table~\ref{tablelabelleddatasets}.

Moreover, in Table~\ref{tablelabelleddatasets}, we label and present the distribution of these taxonomy categories for the 31 studies published between 2012 and 2022, aiming to enhance readability and facilitate efficient comparison. Each row in the Table~\ref{tablelabelleddatasets} represents an individual study, and each column corresponds to the labels assigned in Table~\ref{tabletaxonomy}. This tabulation enables researchers to swiftly identify patterns and trends across studies, offering a condensed yet insightful overview of the sentiments explored in Turkish sentiment analysis. 
With the insights gained from Table~\ref{tablelabelleddatasets}, it becomes apparent that the concept-based approach is infrequently used in these studies, with a prevalent focus on regular type opinion detection which is O1. The concept-based approach, involving a more nuanced understanding of sentiments, is overshadowed by the simplicity and effectiveness of regular-type opinion detection. This nuanced comparison provides a clear overview of the frequency of sentiment analysis tags in studies and enables quick identification and comparison of sentiment analysis studies in Turkish between 2012 and 2022. Such clarity is instrumental for researchers aiming to identify research gaps and focus on areas requiring further exploration within the field.

\subsection{Datasets}
\label{sec:datasets}

To comprehensively explore sentiment analysis studies in the Turkish language, a thorough investigation was conducted on research papers related to `sentiment analysis' and `Turkish dataset' indexed on Scopus between 2012 and 2022\footnote{Search within Article Title, Abstract, Keywords: `sentiment analysis' AND `Turkish dataset'; Date range: 2012-2022; articles indexed in the Scopus database.}. This search resulted in the identification of 31 relevant studies. 

\begin{table}[h!]
    \centering
    \caption{\textit{Studies labelled with sentiment analysis taxonomy classification}}
    \label{tablelabelleddatasets}
    \begin{adjustbox}{max width=1.05\textwidth}
    \begin{tabular}{>{\raggedright\arraybackslash}p{3.5cm}p{0.5cm}p{0.5cm}p{0.5cm}p{0.5cm}p{0.5cm}p{0.5cm}p{0.5cm}p{0.5cm}p{0.5cm}p{0.5cm}p{0.5cm}p{0.5cm}p{0.5cm}p{0.5cm}p{0.5cm}p{0.5cm}p{0.5cm}p{0.5cm}p{0.5cm}p{0.5cm}p{0.5cm}p{0.5cm}p{0.5cm}p{0.5cm}p{0.5cm}p{0.5cm}}
        \toprule 
        \textbf{Studies} & \textbf{O1} & \textbf{O2} & \textbf{E1} & \textbf{E2} & \textbf{E3} & \textbf{E4} & \textbf{E5} & \textbf{T1} & \textbf{T2} & \textbf{T3} & \textbf{D1} & \textbf{D2} & \textbf{G1} & \textbf{G2} & \textbf{G3} & \textbf{G4} & \textbf{G5} & \textbf{G6} & \textbf{A1} & \textbf{A2} & \textbf{A3} & \textbf{A4} & \textbf{A5} & \textbf{A6} & \textbf{A7} & \textbf{A8} \\
        \midrule 
        \citet{Demirtas2013} & X & & X & X & & X & X & & X & & X & X & & X & & & & & & & & & & & X & \\
        \citet{Turkmenoglu2014} & X & & X & X & & X & X & & X & & X & X & & X & & & & & X & & & & & & X & \\
        \citet{Coban2015} & X & & X & X & & X & X & & X & & X & & & X & & & & & & X & & & & & \\
        \citet{Parlar2016} & X & & X & X & & X & & & X & & X & X & & X & & & & & X & & & & & & X & \\
        \citet{Ucan2016} & X & & X & & & X & X & & X & & X & & & X & & & & & X & & & & & & X & \\
        \citet{Hayran2017} & X & & X & X & & X & & & X & & X & X & X & & & & & & X & & & & & \\
        \citet{Mulki2018} & X & & X & & & X & X & & X & & X & X & & X & & & & & & & & & & & X & \\
        \citet{Yuksel2018} & X & & X & & & & X & & X & & X & X & & X & & & X & & X & & & & & \\
        \citet{Amasyali2018} & & X & X & & & X & X & & X & & X & X & & & & & X & & & & X & & & & \\
        \citet{Coban2018} & X & & X & & X & X & X & & X & & X & X & & & & & X & & & & & & & X & & \\
        \citet{Ogul2019} & X & & X & & & & X & & X & & X & & & & & & X & & & X & & & & & & \\
        \citet{Akin2019} & X & & X & & & X & X & & X & & X & & & X & & & X & & & & & & & & X & \\
        \citet{Rumelli2019} & X & & X & X & & X & X & & X & & X & & & X & & & & & X & & & & & & X & \\
        \citet{Shehu2019} & X & & X & X & & X & X & & X & & X & & & X & & & & & X & & & & & & X \\
        \citet{Ersahin2019} & X & & X & & & X & X & & X & & X & & & & & & X & & X & & & & & & X & \\
        \citet{Guven2020} & X & & X & & X & X & X & & & X & X & X & & & & & X & & & X & & & & & X & \\
        \citet{Acikalin2020} & X & & X & X & & X & X & & X & & X & X & & X & & & & & & & & & & X & \\
        \citet{Alqaraleh2021} & X & & X & X & & X & X & & X & & X & & & X & & & & & & & & & & & X & \\
        \citet{Eker2021} & X & & X & X & X & X & X & & X & X & X & X & & X & & & & & & & & & & & X & \\
        \citet{Salur2021} & X & & X & X & & X & X & & X & & X & & & & & & & & & X & & & & &  \\
        \citet{Aydin2021} & X & & X & X & X & X & & & X & & X & & & X & & & X & X & & X & & & & X & X & X \\
        \citet{Zeybek2021} & X & & X & X & X & X & & & X & & X & & & & & & X & & X & & & & & & & X \\
        \citet{Koksal2021} & X & & X & X & & X & X & & X & & & X & & X & & & & & & & & & & & & X \\
        \citet{Aygun2021} & X & & X & & X & X & X & & X & & X & & & & & & X & & X & & & & & X \\
        \citet{Aydogan2023} & X & & X & & & X & X & & X & & X & & & & & & X & & X & & & & & X & X & \\
        \citet{Deniz2022} & X & & X & & & X & X & & & & X & & & & & & X & & & & & & & X & \\
        \citet{Balli2022} & X & & X & X & & & & & X & & X & X & & X & & & & & & & & & & & X &\\
        \citet{Mutlu2022} & X & & X & X & X & X & X & & X & & X & X & & X & & & & & & & & & & X & \\
        \citet{Kabakus2022} & X & & X & X & X & X & X & & X & & X & X & & X & & & X & & X & & & & & & X & \\
        \citet{Alnahas2022}) & X & & X & X & X & x & X & & X & & X & & & & & & X & & & & & & & & X & \\
        \citet{Karayigit2022}) & X & & X & X & X & x & X & & X & & X & & & & & & X & & X & & & & & X & X & \\
        \bottomrule 
    \end{tabular}
    \end{adjustbox}
\end{table}

Subsequently, 23 distinct datasets were collected from publicly available sources and through email requests, and the compiled data were organized within a GitHub repository\footnote{\url{https://github.com/sevvalckc/Turkish-SAD}}. Table~\ref{tablesummaryinfoofdatasets} presents a comprehensive summary of identified studies, providing details such as Article Name, Dataset Name, Data Source, Classification Format, Document Type, Items, Best-Reported Model, Metric, and Metric Score.
In essence, Table~\ref{tablesummaryinfoofdatasets} serves as a comprehensive guide, offering a detailed snapshot of each sentiment analysis study included in our research. It is not only facilitates a nuanced understanding of the methodologies and datasets employed but also provides a foundation for future comparative analyses and the identification of trends within the Turkish sentiment analysis domain.

While our investigation encompasses various datasets, our study particularly focuses on four datasets that were recurrently utilized in multiple publications in the Turkish language. Brief descriptions, contents, and classification types of five datasets are as follows:
The Turkish Movie Reviews Dataset (TMRD), and The Turkish Multidomain Products Reviews (TMPR), were prepared by \citet{Demirtas2013}. Notably, these datasets have found application in three additional studies by \citet{Parlar2016}, \citet{Mulki2018}, and \citet{Alqaraleh2021}. TMRD encompasses comments and reviews gathered from beyazperde.com, while TMPR consists of comments and reviews collected from hepsiburada.com. Both datasets are tailored for binary classification, with TMRD comprising 5,331 positive and 5,331 negative instances, and TMPR presenting an equal distribution of 2,800 positive and 2,800 negative instances. We predict instances of these two datasets using these four models. The datasets were combined into a single dataset, referred to as TMRD-TMPR in our paper.

In addition to the aforementioned datasets, our study also incorporates the Twitter Dataset prepared by \citet{Turkmenoglu2014}. The Twitter Dataset is composed of tweets collected from Twitter and adhere to a binary classification format and includes 1,677 positive and 1,301 negative instances. \citet{Turkmenoglu2014} highlighted the effectiveness of Support Vector Machines (SVM) in the dataset, achieving an accuracy of .85 for the Twitter Dataset. Notably, \citet{Aydin2021} utilized the same dataset in their study.

\begin{table}[H]
    \centering
    \caption{Summary information of datasets, the best performing model, and accuracy score according to the cited paper are reported.}
    \label{tablesummaryinfoofdatasets}
    \begin{adjustbox}{max width=\textwidth}
    \begin{tabular}{>{\raggedright\arraybackslash}p{2cm} p{3cm} p{3cm} p{3cm} p{1cm} p{2cm} p{3cm} p{2cm} p{2cm}}
        \toprule
        \textbf{Referans} & \textbf{Dataset Name} & \textbf{Data Source} & \textbf{Document Type} & \textbf{CF} & \textbf{Items} & \textbf{Best Reported Model} & \textbf{Metric} & \textbf{Metric Score} \\
        \midrule
        \citet{Demirtas2013} & Turkish Movie Reviews Dataset & Beyazperde.com & Comment & BC & Pos: 5,331 Neg: 5,331 & & & \\
        & Turkish Multidomain Products Reviews & Hepsiburada.com & Comment & BC & Pos: 2,800 Neg: 2,800 & & & \\
        \citet{Turkmenoglu2014} & Twitter Dataset & Twitter & Tweet & BC & Pos: 1,677 Neg: 1,301 & SVM & Accuracy & .85 \\
        & Movie Dataset & Beyazperde.com & Comment & BC & Pos: 13,224 Neg: 7,020 & SVM & Accuracy & .895 \\
        \citet{Coban2015}) & Twt & Twitter & Tweet & BC & Pos: 1,000 Neg: 1,000 & MNB & Accuracy & .66 \\
        \citet{Parlar2016} & Turkish Multidomain Products Reviews & Hepsiburada.com & Comment & BC & Pos: 2,800 Neg: 2,800 & NBM & F-Score & .8042 \\
        \citet{Ucan2016} & Movie Review & Beyazperde.com & Comment & BC & Pos: 26,700 Neg: 26,700 & TSDp & Accuracy & .807 \\
        & Hotel Review & Otelpuan.com & Comment & BC & Pos: 5,800 Neg: 5,800 & SWN & Accuracy & .846 \\
        \citet{Hayran2017} & Turkish Sentiment Dataset & Twitter & Tweet & BC & Pos: 16,000 Neg: 16,000 & Dvot & Accuracy & .8005 \\
        \citet{Mulki2018} & Turkish Movie Reviews Dataset & Beyazperde.com & Comment & BC & Pos: 5,331 Neg: 5,331 & NB & Accuracy & .928 \\
        & Turkish Multidomain Products Reviews & Hepsiburada.com & Comment & BC & Pos: 2,800 Neg: 2,800 & NB & Accuracy & Kitchen: .85 DvD: .814 Electronic: .857 Books: .886 \\
        \citet{Yuksel2018} & Foursquare Venue and Venue Comments Data & Foursquare & Comment & BC & Total: 7,086 & RT-SKDA & Accuracy & .8197 \\
        \citet{Amasyali2018} & Twitter Dataset & Twitter & Tweet & TC & Pos: 4,579 Neg: 6,886 Neu: 5,822 & SDG & Success & .6925 \\
        \citet{Coban2018} & VS1 & Twitter & Tweet & TC & Pos: 756 Neg: 1,287 Neu: 957 & BoW & Accuracy & .5384 \\
        & VS2 (Turkish Sentiment Dataset) & Twitter & Tweet & BC & Pos: 16,000 Neg: 16,000 & BoW & Accuracy & .7221 \\
        \citet{Ogul2019}) & VS2 & Twitter & Tweet & TC & Pos: 4,579 Neg: 6,886 Neu: 5,822 & SMOTE & Accuracy & .6859 \\
        \citet{Akin2019} & Restaurant Dataset & Online & Comment & BC & Total: 756 & with-TL & Accuracy & .901 \\
        & Product Dataset & Online & Comment & BC & Pos: 220,284 Neg: 14,881 & with-TL & Accuracy & .834 \\
        \citet{Rumelli2019} & Sentiment Analysis Twitter Dataset & Hepsiburada.com & Comment & BC & Pos: 221,071 Neg: 13,012 & kNN & Accuracy & .738 \\
        \citet{Shehu2019} & Dataset & Twitter & Tweet & TC & Total: 3,000 and 10,500 & RF & Accuracy & .885 \\
        \citet{Ersahin2019} & Movie Review & Beyazperde.com & Comment & BC & Pos: 26,700 Neg: 26,700 & Hybrid (SVM+eSTN) & Accuracy & .8631 \\
        & Hotel Review & Otelpuan.com & Comment & BC & Pos: 5,800 Neg: 5,800 & Hybrid (SVM+eSTN) & Accuracy & .9196 \\
        & Tweets Dataset & Twitter & Tweet & BC & Total: 1,756 & Hybrid (NB+eSTN) & Accuracy & .8337 \\
    \bottomrule
    \end{tabular}
    \end{adjustbox}
    \caption*{Note: CF = Classification Format, BC = Binary Classification, TC = Tentary Classification, EC = Emotion Classification, Pos = Positive, Neg = Negative, Neu = Neutral.}
\end{table}

\newpage 

\begin{table}[H]
    \ContinuedFloat
    \centering
    \caption{(continued) Summary information of datasets, the best performing model, and accuracy score according to the cited paper are reported.}
    \begin{adjustbox}{max width=\textwidth}
    \begin{tabular}{>{\raggedright\arraybackslash}p{2cm} p{3cm} p{3cm} p{3cm} p{1cm} p{2cm} p{3cm} p{2cm} p{2cm}}
        \toprule
        \textbf{Referans} & \textbf{Dataset Name} & \textbf{Data Source} & \textbf{Document Type} & \textbf{CF} & \textbf{Items} & \textbf{Best Reported Model} & \textbf{Metric} & \textbf{Metric Score} \\
        \midrule
        \citet{Guven2020} & Dataset & Twitter & Tweet & EC & Happy: 800 Sad: 800 Shocked: 800 Fear: 800 Angry: 800 & GAA & Accuracy & .87 \\
        \citet{Acikalin2020} & Movie Review & Beyazperde.com & Comment & BC & Pos: 26,700 Neg: 26,700 & BERT & Accuracy & .9134 \\
        & Hotel Review & Otelpuan.com & Comment & BC & Pos: 5,800 Neg: 5,800 & fastText-EN & Accuracy & .8080 \\
        \citet{Alqaraleh2021} & Turkish Movie Reviews & Beyazperde web pages & Comment & BC & Total: 34,990 & SA system & Accuracy & .8329 \\
        \citet{Eker2021} & Dataset & Twitter & Tweet & EC & Happy: 800 Sad: 800 Shocked: 800 Fear: 800 Angry: 800 & BiLSTM-RNN & Accuracy & .952 \\
        \citet{Salur2021} & Turkish Tourism ABSA Dataset & TripAdvisor & Comment & TC & & & & \\
        \citet{Aydin2021} & Movie Dataset & Beyazperde.com & Comment & BC & Pos: 13,224 Neg: 7,020 & SVM & Accuracy & .9098 \\
        & Twitter Dataset & Twitter & Tweet & BC & Pos: 743 Neg: 973 & SVM & Accuracy & .7964 \\
        \citet{Zeybek2021} & Turkish Movie Reviews Dataset & Beyazperde.com & Comment & BC & Pos: 5,331 Neg: 5,331 & RNTN & Accuracy & .8984 \\
        & Turkish Multidomain Products Reviews & Hepsiburada.com & Comment & BC & Pos: 2,800 Neg: 2,800 & RNTN & Accuracy & Books: .8608 DVD: .8295 Electr.8295 Electr.: .8666 Kitchen: .8173 \\
        \citet{Koksal2021} & BounTi & Twitter & Tweet & TC & Total: 7,964 & BERTurk & Accuracy & .0745 \\
        \citet{Aygun2021} & Turkish Data from Banda Dataset & Twitter & Tweet & BC & Total: 928,402 & BERTurk & Accuracy & .89 \\
        \citet{Aydogan2023} & TRSAv1 & Online shopping sites & Comment & TC & Total: 150,000 (50,000 per neutral, positive, negative) & B-LSTM with GloVe & Accuracy & .8361 \\
        \citet{Deniz2022} & Dataset & Trendyol & Comment &  & Total: 51,394 & & microPrecision & .9157 \\
        \citet{Balli2022} & SentimentSet Dataset & Twitter & Tweet & TC & Total: 2,551 & LSTM & Accuracy & .86 \\
        & Public Dataset & Twitter & Tweet & BC & Total: 11,000 & LSTM & Accuracy & .86 \\
        \citet{Mutlu2022} & Twitter Dataset & Twitter & Tweet & TC & 3,952 & T-BERTmarked-MP & F-1 Score & .669 \\
        \citet{Kabakus2022} & COVID-19 Dataset & Twitter & Tweet & TC & Pos: 3,290 Neg: 852 Neu: 45,802 & Proposed Model & Accuracy & .952 \\
        \citet{Alnahas2022} & Dataset & E-Commerce Websites & Comment & TC & Total: 602,202 & Suggested Ensemble Model & F-1 Score & .847 \\
        \citet{Karayigit2022} & Dataset1 & Instagram & Comment & TC & Pos: 671 Neg: 1,473 Neu: 743 & BERT-based transfer learning model & Accuracy & .7864 \\
        & Dataset2 & Instagram & Comment & TC & Pos: 2,074 Neg: 1,564 Neu: 3,183 & BERT-based transfer learning model & Accuracy & .7120 \\
        \bottomrule
    \end{tabular}
    \end{adjustbox}
    \caption*{Note: CF = Classification Format, BC = Binary Classification, TC = Tentary Classification, EC = Emotion Classification, Pos = Positive, Neg = Negative, Neu = Neutral.}
\end{table}

Furthermore, our study integrates the Twt dataset, prepared by \citet{Coban2015}. Comprising 2,000 items, equally distributed with 1,000 positive and 1,000 negative instances, the Twt dataset is a unique addition to our analysis. This dataset was collected by \citet{Coban2015} using the Twitter API service. It is important to note that the Twt dataset, unlike others, lacks public access and has not been assigned a formal name. For our study, we will refer to it as the ``Twt" dataset. The categorization of tweets as positive or negative in this dataset relies solely on emoticons, where ``:)", ``:D," and similar expressions denote positivity, while ``:(", ``=(", ``;(" signify negativity. Following preprocessing steps, including the removal of hashtags, URLs, and usernames, the data were randomly collected without being constrained to a specific subject. Due to the temporal constraints of Twitter's character limit at the time of data collection, posts were limited to 140 characters. \citet{Coban2015} highlighted the effectiveness of Multinomial Naive Bayes (MNB) in Twt dataset, achieving an accuracy of .66 showcasing its unique characteristics and challenges. Furthermore, \citet{Kabakus2022} utilized the same Twt dataset in their study. 

Similarly, we will provide additional details about another dataset, referred to briefly as ``Humir" as introduced by \citet{Ucan2016}. The dataset acquisition process involved the selection of two widely accessed platforms, namely ``beyazperde.com" for movie reviews and ``otelpuan.com" for hotel reviews. A total of 53,400 movie reviews were collected, with an average length of 33 words. These reviews were rated by the authors on a scale of 1 to 5 stars, ensuring a balanced representation of both positive and negative sentiments. Similarly, for hotel reviews, a set of 11,600 positive and negative reviews were chosen from 18,478 reviews extracted from 550 hotels. The average length of these hotel reviews was 74 words, more than double that of the movie reviews. This dataset's versatility is further underscored by its utilization in the study conducted by \citet{Acikalin2020} emphasizing its relevance across different analyses within the realm of sentiment analysis.

\section{Methodology}
\label{sec:methodology}

In our study, we have evaluated four state-of-the-art models to study sentiment in various texts, with a special focus on Turkish. These models are XLM-T, BERTurk (enhanced with the BounTi dataset), the TSAM available on Hugging Face, and TurkishBERTweet developed by ViralLab. These models predicts one of positive, negative, and neutral for a given text.

\begin{itemize}
    \item XLM-T\footnote{\url{https://huggingface.co/cardiffnlp/twitter-xlm-roberta-base-sentiment}}: This model enables us to analyze sentiments across various languages, crucial for understanding multilingual sentiment on platforms like Twitter. Model is a multilingual XLM-RoBERTa-based model trained on approximately 198M tweets for sentiment analysis. 
\end{itemize}
\begin{itemize}
    \item BERTurk with the BounTi Dataset\footnote{\url{https://github.com/boun-tabi/BounTi-Turkish-Sentiment-Analysis}}: Specifically a fine-tuned version of the pre-trained BERTurk model, when combined with the BounTi dataset, offers us a valuable asset of sentiment prediction tool for Turkish texts. The BounTi dataset consists of Turkish tweets \citet{Koksal2021}. 
\end{itemize}
\begin{itemize}
    \item TSAM on Hugging Face\footnote{\url{https://huggingface.co/emre/turkish-sentiment-analysis}}: This model is optimized for Turkish sentiment analysis, utilizing a diverse set of Turkish test texts for training. It is designed to accurately detect different expressions of sentiment. 
\end{itemize}
\begin{itemize}
    \item TurkishBERTweet by ViralLab\footnote{\url{https://huggingface.co/VRLLab/TurkishBERTweet\#-turkishbertweet-fast-and-reliable-large-language-model-for-social-media-analysis}}: This model excels in analyzing sentiments expressed in Turkish tweets. Its training on 894 million Turkish tweets allows it to decode sentiments, casual expressions, emojis, and informal language found in social media texts. Model is a large language model based on RoBERTa architecture.
\end{itemize}

Through the evaluation of four models, provide a comprehensive evaluation of sentiment analysis models. This includes assessing each model's applicability to Turkish sentiment datasets. We also compile studies on sentiment analysis and cross-dataset classification to provide detailed dataset information and compare their similarities. This approach enables us to assess the strengths and potential limitations of each dataset and model comprehensively. Evaluation of various models and datasets facilitates a detailed exploration of sentiment analysis, contributing valuable insights to the domain of Turkish NLP studies.

\section{Results}
\label{sec:results}
The results we have obtained shed light the performance of the XLM-T, TSAM, TurkishBERTweet, and BERTurk models across various datasets. To enable more detailed analysis and facilitate the replication of our findings, we have made our code and models publicly available in our repository\footnote{\url{https://github.com/sevvalckc/Turkish-SAD}}. This resource includes scripts for training the models, making predictions, and calculating key metrics such as Accuracy, Recall, Precision, and F1 scores. Our repository also offers a step-by-step guide, ensuring that both researchers and practitioners can easily leverage our work to benchmark or expand upon the sentiment analysis for Turkish texts.

In Table~\ref{tableresult}, we report detailed performance of these models using the aforementioned metrics. It is important to note that all models were designed to generate predictions based on a multi-classification format. However, the datasets vary, with some being tailored for multi-classification and others for binary classification. To address this variation and clarify the performance across different formats, Table~\ref{tableresult} includes scores calculated without modifying the classification formats of either the datasets or the models. Additionally, it presents scores recalculated for both datasets and models in a binary classification format, achieved by excluding ``neutral" labels.

\begin{table}[h!]
    \centering
    \caption{\textit{Model performance on different datasets.}}
    \label{tableresult}
    \begin{adjustbox}{max width=\textwidth}
    \begin{tabular}{>{\raggedright\arraybackslash}p{3cm} p{3cm} p{3cm} p{1.5cm} p{1.5cm} p{1.5cm} p{1.5cm} p{1.5cm} p{1.5cm} p{1.5cm}}
        \toprule
        \textbf{Dataset Name} & \textbf{Model} & \textbf{Classification} & \textbf{Accuracy} & \textbf{Recall Micro} & \textbf{Recall Macro} & \textbf{Precision Micro} & \textbf{Precision Macro} & \textbf{F1 Micro} & \textbf{F1 Macro} \\
        \midrule
        Twt & XLM-T & Model: Multi, Dataset: Binary & .83 & .83 & .55 & .83 & .63 & .83 & .59 \\
        & & Both: Binary & .92 & .95 & .94 & .95 & .95 & .95 & .95 \\
        & TSAM & Model: Multi, Dataset: Binary & .59 & .59 & .44 & .59 & .39 & .59 & .36 \\
        & & Both: Binary & .59 & .59 & .66 & .59 & .59 & .59 & .54 \\
        & TurkishBERTweet & Model: Multi, Dataset: Binary & .57 & .57 & .58 & .57 & .38 & .57 & .45 \\
        & & Both: Binary & .86 & .86 & .87 & .86 & .85 & .86 & .86 \\
        & BERTurk & Model: Multi, Dataset: Binary & .56 & .56 & .52 & .56 & .37 & .56 & .40 \\
        & & Both: Binary & .75 & .75 & .78 & .75 & .71 & .75 & .72 \\
        \midrule
        Turkmenoglu & XLM-T & Both: Multi & .61 & .61 & .61 & .61 & .62 & .61 & .61 \\
        & & Both: Binary & .82 & .82 & .82 & .82 & .81 & .82 & .81 \\
        & TSAM & Both: Multi & .47 & .47 & .53 & .47 & .43 & .47 & .35 \\
        & & Both: Binary & .68 & .68 & .75 & .68 & .64 & .68 & .63 \\
        & TurkishBERTweet & Both: Multi & .56 & .56 & .62 & .56 & .58 & .56 & .56 \\
        & & Both: Binary & .82 & .82 & .83 & .82 & .82 & .82 & .82 \\
        & BERTurk & Both: Multi & .61 & .61 & .62 & .61 & .62 & .61 & .61 \\
        & & Both: Binary & .81 & .81 & .81 & .81 & .81 & .81 & .81 \\ 
        \midrule
        TMRD TMPR & XLM-T & Model: Multi, Dataset: Binary & .56 & .56 & .48 & .56 & .37 & .56 & .42 \\
        & & Both: Binary & .72 & .72 & .72 & .72 & .72 & .72 & .72 \\
        & TSAM & Model: Multi, Dataset: Binary & .73 & .73 & .52 & .73 & .49 & .73 & .48 \\
        & & Both: Binary & .74 & .74 & .78 & .74 & .74 & .74 & .73 \\
        & TurkishBERTweet & Model: Multi, Dataset: Binary & .46 & .46 & .49 & .46 & .31 & .46 & .37 \\
        & & Both: Binary & .73 & .73 & .73 & .73 & .73 & .73 & .73 \\
        & BERTurk & Model: Multi, Dataset: Binary & .58 & .58 & .47 & .58 & .38 & .58 & .41 \\
        & & Both: Binary & .69 & .69 & .70 & .69 & .69 & .69 & .68 \\
        \midrule
        Humir & XLM-T & Model: Multi, Dataset: Binary & .81 & .81 & .56 & .81 & .54 & .81 & .55 \\
        & & Both: Binary & .85 & .85 & .85 & .85 & .85 & .85 & .85 \\
        & TSAM & Model: Multi, Dataset: Binary & .97 & .97 & .65 & .97 & .65 & .97 & .65 \\
        & & Both: Binary & .97 & .97 & .97 & .97 & .97 & .97 & .97 \\
        & TurkishBERTweet & Model: Multi, Dataset: Binary & .64 & .64 & .54 & .64 & .42 & .64 & .47 \\
        & & Both: Binary & .80 & .80 & .81 & .80 & .78 & .80 & .79 \\
        & BERTurk & Model: Multi, Dataset: Binary & .83 & .83 & .57 & .83 & .55 & .83 & .56 \\
        & & Both: Binary & .86 & .86 & .86 & .86 & .86 & .86 & .86 \\
        \bottomrule
    \end{tabular}
    \end{adjustbox}
\end{table}

In binary classification tasks, where the objective was to differentiate between positive and negative sentiments, XLM-T and BERTurk showcased superior performance across the board. Notably, XLM-T achieved a remarkable accuracy and F1 score of .92 and .95, respectively, on the Twt dataset, highlighting its proficiency in binary sentiment analysis. Similarly, BERTurk exhibited strong performance with an accuracy of .86 and F1 score of .86 on the Twt dataset, further cementing its reliability in binary classification contexts.

TSAM, while showing moderate performance on the Twt dataset with accuracy and F1 scores of .59 and .54, respectively, demonstrated exceptional strength on the Humir dataset with near-perfect accuracy and F1 scores of .97. This indicates the TSAM's potential effectiveness in specific binary classification scenarios, especially when tuned for particular types of data.

TurkishBERTweet, achieving an accuracy of .86 and an F1 score of .86 on the Twt dataset, also illustrated commendable binary classification capabilities, showcasing its potential utility in sentiment analysis tasks that require distinguishing between positive and negative sentiments.

Multi-class classification tasks, which included the challenge of identifying neutral sentiments alongside positive and negative ones, revealed a different aspect of model performance. XLM-T showed consistent and promising results, particularly on the Turkmenoglu dataset, with an accuracy of .61 and an F1 score of .61. This underscores XLM-T's versatility and strong adaptability across different classification formats.

BERTurk and TurkishBERTweet displayed moderate success in multi-class scenarios, with BERTurk achieving an accuracy of .61 and an F1 score of .61 on the Turkmenoglu dataset. These outcomes suggest a balanced performance but also room for improvement in handling more nuanced classification tasks.

TSAM encountered challenges in multi-class classification, particularly evidenced by its lower accuracy (.47) and F1 score (.35) on the Turkmenoglu dataset. This performance difference highlights TSAM’s limitations in complex classification scenarios, despite its binary classification strengths.
In scenarios where the dataset and model classification formats did not match, the performance of each model varied, revealing their flexibility and limitations. For instance, XLM-T managed a .83 accuracy in the Twt dataset when applying a multi-class model to a binary classified dataset, showcasing its adaptability. However, the mismatch scenario generally resulted in lower performance metrics across models compared to when the classification formats were aligned, indicating the importance of matching model and dataset formats for optimal performance.

\section{Discussion}
\label{sec:discussion}
In this study, we conducted a comprehensive exploration of sentiment analysis studies in the Turkish language, spanning from data collection to model evaluation. We curated 23 distinct datasets from publicly available sources and through email requests, organizing the compiled data within a GitHub repository. Furthermore, we systematically reviewed 31 relevant studies indexed on Scopus between 2012 and 2022, providing a detailed snapshot of each sentiment analysis study in our research.

The integration of four state-of-the-art models—XLM-T, BERTurk, TSAM, and TurkishBERTweet—allowed us to analyze sentiments in Turkish texts comprehensively. These models were evaluated across various datasets, including the Turkish Movie Reviews Dataset (TMRD), Turkish Multidomain Products Reviews (TMPR), Twt dataset, and Humir dataset, among others. We also utilized a taxonomy to facilitate a nuanced understanding of the methodologies and datasets employed in Turkish sentiment analysis studies.

The evaluation of sentiment analysis models across various datasets provides valuable insights into their performance and suitability for different text corpora. Our study incorporated a diverse range of datasets, including movie reviews, product reviews, Twitter data, and COVID-19 related tweets, among others. Each dataset presents unique characteristics, such as data source, classification format, and size, which significantly impact model performance.

Across binary classification tasks, datasets like the Twt dataset and Turkish Movie Reviews Dataset (TMRD) demonstrated balanced distributions of positive and negative instances, contributing to relatively higher model accuracies. Multi-class classification tasks, which involve identifying neutral sentiments alongside positive and negative ones, revealed additional complexities. However, models like XLM-T exhibited promising adaptability and achieved competitive accuracy scores across multi-class scenarios.

The performance of sentiment analysis models varied across datasets, reflecting the interplay between model architectures, training strategies, and dataset characteristics. XLM-T emerged as a top performer, showcasing superior accuracy and F1 scores in binary classification tasks across multiple datasets. This can be attributed to XLM-T's cross-lingual learning capabilities, enabling it to effectively capture sentiment nuances in Turkish texts with diverse linguistic features. BERTurk, while demonstrating strong performance overall, exhibited slightly lower accuracies compared to XLM-T in binary classification tasks. However, it showcased competitive results in multi-class scenarios, indicating its robustness in handling nuanced sentiment analysis tasks. The TSAM and TurkishBERTweet also displayed moderate to high accuracies across datasets, highlighting their effectiveness in capturing sentiment expressions in Turkish texts.

The relationship between model performance and dataset characteristics is complex and multifaceted. Generally, models performed better on datasets with balanced class distributions, where positive and negative instances were evenly represented. This suggests that the measurement of the model accuracy is influenced by the availability of sufficient evaluation data for each class.

Moreover, the linguistic characteristics of the dataset, such as informal language, emoticons, and domain-specific jargon, significantly impact model performance. Models trained on datasets with diverse linguistic features, like social media texts, demonstrated higher adaptability and robustness in capturing sentiment nuances.

The discrepancy in model performance across datasets underscores the importance of dataset characteristics and preprocessing in sentiment analysis studies. Researchers must carefully curate datasets that align with the target application domain and ensure adequate representation of sentiment expressions to improve model generalization and performance.

The performance scores reported in original studies have been significantly updated in our study through the application of advanced deep learning models. Traditional machine learning methods, such as Multinomial Naive Bayes (MNB) and Support Vector Machines (SVM), achieved notable results in earlier research but were constrained by their limited ability to handle linguistic complexity. For instance, \citet{Coban2015} reported an accuracy of 0.66 on the Twt dataset using MNB. In contrast, our deep learning models have considerably improved upon these results, with XLM-T achieving an accuracy of 0.92, TurkishBERTweet reaching 0.86, and BERTurk attaining 0.75. These advancements underscore the enhanced capabilities of modern models to capture linguistic nuances and domain-specific variations, demonstrating how the integration of deep learning can substantially elevate the benchmarks established by traditional approaches.

The quality and diversity of datasets also have a significant impact on model performance. In tasks like sentiment analysis, factors such as imbalanced class distributions, insufficient sample sizes, and noisy data can negatively affect the model's ability to generalize. The experiments conducted on different datasets in this study reveal that the performance of the models varies according to the characteristics of the dataset. This situation highlights the critical importance of dataset preparation and preprocessing in sentiment analysis studies.

Lastly, the linguistic features of Turkish, such as its agglutinative structure and rich inflectional systems, present specific challenges for natural language processing models. This study has shown that XLM-T is more effective than BERTurk in overcoming these types of challenges. These findings can guide future work and contribute to the development of more successful natural language processing models for languages with complex structures.

\section{Conclusions and Future Work}
\label{sec:conclusions_and_future_work}
This study underscores the importance of model selection and dataset preparation in the field of Turkish sentiment analysis, highlighting how model architecture, data characteristics, and classification alignment contribute significantly to model performance. Our findings demonstrate that models like XLM-T and BERTurk, when applied to diverse datasets, achieve robust results in both binary and multi-class classification tasks, emphasizing the need for dataset-model compatibility. Dataset curation and preprocessing also play a critical role, as balanced class distributions and domain-specific data improve model adaptability and accuracy. These insights are crucial for sentiment analysis studies, where adequate representation of sentiment classes greatly impacts model success.

While significant advancements have been made using machine learning and deep learning approaches, concept-based sentiment analysis studies remain absent in the Turkish sentiment analysis landscape. These studies, which often utilize semantic networks or ontologies to analyze text, could provide valuable insights into the contextual and relational aspects of sentiment, especially in capturing complex or implicit sentiments. The integration of such methods could further enrich the analytical depth and application scope of sentiment analysis in Turkish.

Future work can build on these findings by creating more advanced models and refining data processing methods. For instance, using larger and more varied datasets could make models more adaptable to different types of sentiment. Additionally, developing benchmark datasets and standardized evaluation metrics for Turkish sentiment analysis would make it easier to compare results across studies, helping establish common standards in the field. Techniques like transfer learning or few-shot learning could also be explored to improve model performance when data is limited, ultimately advancing the effectiveness of sentiment analysis in Turkish.

\bibliographystyle{plainnat}
\bibliography{references}

\end{document}